# Metamorphic Detection of Adversarial Examples in Deep Learning Models With Affine Transformations

Rohan Reddy Mekala    Gudjon Einar Magnusson    Adam Porter    Mikael Lindvall    Madeline Diep

Fraunhofer CESE

{rreddy|gmagnusson|aporter|mikli|mdiep}@fc-md.umd.edu

5700 Rivertech Court #210, Riverdale, MD 20737, USA

*Abstract—* **Adversarial attacks are small, carefully crafted perturbations, imperceptible to the naked eye; that when added to an image cause deep learning models to misclassify the image with potentially detrimental outcomes. With the rise of artificial intelligence models in consumer safety and security intensive industries such as self-driving cars, camera surveillance and face recognition, there is a growing need for guarding against adversarial attacks. In this paper, we present an approach that uses metamorphic testing principles to automatically detect such adversarial attacks. The approach can detect image manipulations that are so small, that they are impossible to detect by a human through visual inspection. By applying metamorphic relations based on distance ratio preserving affine image transformations which compare the behavior of the original and transformed image; we show that our proposed approach can determine whether or not the input image is adversarial with a high degree of accuracy.**

## I. Introduction

Deep learning with neural networks [1] has become a very important facet of modern engineering involving image processing. The ability of neural networks to achieve high accuracy on complex tasks has led to widespread acceptance in consumer-facing industries. From self-driving cars to smart surveillance cameras, we are seeing a gradual shift from manual control towards using trained neural networks [2]. However, the robustness and security of these methods have been called into question. Carefully crafted perturbations (referred to as adversarial perturbations) in these images have been known to break the best of machine learning models with a worrying degree of accuracy [3][5]. Fig. 1 illustrates how an airplane was classified as a parachute with a high level of confidence when the imperceptible adversarial noise gets added to individual pixels. Furthermore, it is possible for an external attacker with no knowledge of the in-house model to develop perturbation attacks targeted towards misclassifying any object as a rogue label on the in-house model [4].

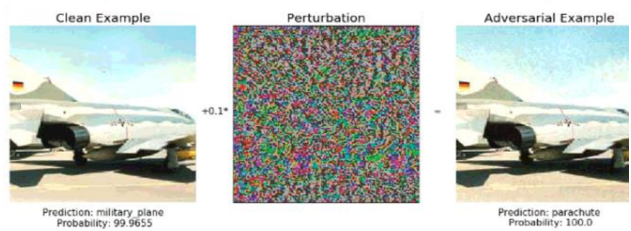

**Figure 1: Adversarial Attack with Epsilon of 0.1**

Popular state-of-the-art object detection algorithms like YOLO[6] have been successfully attacked with adversarial examples [7]. This is the source of the much-publicized example where a stop sign was misclassified by a moving car, just by adding tiny stickers containing the perturbation needed to distort the accuracy of the network [7]. Even commercial grade artificial intelligence models like the Google Cloud Vision API were successfully broken with black box threat models by Ilyas et al [8], causing images of a 3D printed turtle to be misclassified as a rifle, regardless of camera angle. These examples show that adversarial examples can be used in the physical world to attack deep learning systems.

In this paper we present an idea for detecting adversarial attacks at runtime. This type of technique would be of great value for safety critical systems that use deep learning for decision making. The US NAVY, which partially funded this work, is a case in point because they have expressed interest in using deep learning for various image classification tasks involving territory surveillance. Our suggested approach aims to reduce the risk of adversarial attacks by verifying the authenticity of the input before making critical decisions.

Our proposed approach is based on the hypothesis that adversarial perturbations in images can be efficiently detected by comparing the classification of an image with affine transformation of the same image. The idea is based on metamorphic testing principles [21] in that the approach should automatically, be able to detect such perturbations without needing to know the true output label for our test images (clean or adversarial). Additionally, we do not need to implement a machine learning model to capture the adversarial behavior of an image. For example, let us assume that the test image represents a navy ship and the transformation under use is rotation. If the trained neural network classifies the object in the image as a navy ship, then the same neural network should classify the rotated copy of the image as a navy ship as well (with little to no change in the prediction confidence). If the classification of the original and the transformed images are different and in accordance with the rules defined as part of our metamorphic relations, then, we conclude that the image has been manipulated. If our hypothesis is correct, our approach can be used for runtime verification since it can determine on-the-fly whether or not images serving as input to an AI-based system are manipulated. Images that are determined to be manipulated would be discarded and a message would be sent to the commander of the ship that the AI-based system might be compromised.

## II. BACKGROUND

There are currently two prominent areas of research on detecting adversarial images. a) **Creating a neural network to differentiate adversarial images from actual images** – This involves creating a new model which is trained on adversarial and clean examples labeled as their respective classes; thus being able to detect adversarial examples on test data. Methods in this category require large set of training data and still have a propensity to overfit. Additionally, they are computational- and time-intensive due to the complexity of the network and dimensionality of input images. b) **Multi model prediction** – These methods tend to train multiple models and detect adversarial examples under the assumption that the same example would be improbable to fool multiple models with consistency. However, this assumption may not hold. It has been shown that many adversarial attacks do a good job of generalizing between different models [8].

The current solutions and testing procedures for defense against adversarial examples on the program under test (the detection model) (PUT) [9] have been dodgy at large due to the high dimensionality in the spatial space of adversarial permutations. As a consequence, they do not capture quantifiable behavior in adversarial perturbations which could be used to create systematic testing procedures for adversarial detection. This has crippled the ability of machine learning algorithms in being able to prevent attacks with an accountable degree of accuracy. The current solutions face the following problems, making the task of a tester finding adversarial examples extremely difficult:

1. Extensive use of computational resources to creating complex neural networks capable of capturing the permutations in adversarial noise vectors
2. Voluminous demand for clean and adversarial examples to train and validate the model
3. Highly expensive nature of manual work needed to find the "correct" label for our detection model

The shortcomings of existing solutions inspired us to explore a metamorphic (testing) approach towards finding differentiable properties between adversarial and unperturbed images. Metamorphic testing is a novel testing procedure comprising tuples of inputs and outputs where the "correctness" of a test case is defined by the "degree of change" in the output when applying a transformation to the test input [11]. A conditional rule defining the degree of expected change in the output is defined formally as a Metamorphic Relation (MR). This approach towards testing reduces the need to expend time in formulating testing "oracles" that define formal relations between the input and output tuples.

We hypothesize that a more efficient way to detect adversarial examples would be to find inherent properties in the behavior of these images towards affine linear transformations that differentiate them from regular images. Armed with the knowledge of probabilistic axioms governing these properties, we created metamorphic relations which would help achieve comparable/better levels of detection accuracy while not having to facing the drawbacks of current methods explained earlier. This would help defenders manufacture real-time defenses against adversarial attacks with a short turn-around time. Our training environment was chosen carefully to simulate data distributions and deep learning environments used in security intensive maritime environments. The subsequently trained deep learning model was made robust to image transformations using augmented training data constructed using random image transformations. The model was subsequently subjected to the untargeted Fast Gradient Signed Attack method (FGSM) [16] to generate adversarial examples. Using metamorphic testing principles, we established clear demarcations in the behavior of adversarial and clean images towards the output label confidence/probability when subjected to four affine image transforms: Rotation, Shear, Scaling and Translation.

## III. CONTRIBUTIONS

To the best of our knowledge, our paper advances the work done towards adversarial detection and metamorphic testing in the following novel ways:

- This is the first attempt at using metamorphic testing to move away from using the traditional deep learning/machine learning approach to detect adversarial examples on deep learning-based classifier models involving higher resolution datasets like ImageNet [10] (224x224x3)
- This is the first attempt at using metamorphic testing over deep learning models built using transfer learning on low volume training data
- Based on experimental results, we propose, test and confirm a new iterative generalization method useful in filtering out adversarial anomalies. Results from our tested hypotheses are then converted into probabilistic metamorphic relations, that minimize the constraints from traditional models while requiring no manual work in classifying the "accurate" output label for our test data. Our method achieves adversarial detection accuracies of up to 96.85% on unseen data.

The results show that deep learning models used for high stake security intensive maritime applications would benefit from adopting our proposed metamorphic testing approach. This would not only decrease manual testing time and computation costs (enabling faster response time) defending against adversarial attacks, but would also deliver comparable/better accuracy.

## IV. RELATED WORK

The process of using affine transforms as a defense against adversarial attacks on higher dimensional images is relatively untouched with there being exciting opportunities for future research, as demonstrated in our paper. The work on applications of metamorphic testing in the context of machine learning by Dwarakanath et al. [13] was a huge inspiration in encouraging us to delve further into exploring applications of metamorphic testing over adversarial examples in machine learning models. Additionally, we could find one instance of research by Tian et al. [14], that uses transformative affine procedures for detecting adversarial examples where they test for defense against the

Carlini Wagner Attack [15] with a machine learning approach (combining the classifier and detector into a single model) achieving detection accuracy of 70% on white box attacks. This method was tested on lower dimensional datasets like CIFAR(32x32x3) and MNIST(28x28x1).

Our work focuses on a purely metamorphic approach that does not require machine learning models for the detection model and relies on metamorphic relations comprising an iterative function across incremental degrees/units of transformation used to detect adversarial examples. This significantly helps with solving the oracle problem in testing over deep learning models. Additionally, we show the efficacy of our method over higher dimensional images like in ImageNet (224x224x3) used over transfer learning models. Using the Fast Gradient Sign Attack (FGSM) [16] as a reference adversarial example creation mechanism, we were able to achieve 96.85% detection accuracy with white box attacks on unseen data.

## V. DEEP LEARNING MODEL

### A. Developing a target model

We first simulate a maritime environment with an in-house neural net-based classifier using the ImageNet dataset for our input images. With ~200,000 images and 100+ categories, we decided to constrain our classifier model training to ~11,000 images and 15 categories applicable to maritime environments. The categories for our output labels are: lighthouse, container ship, ocean liner, parachute, military plane, army tank, pirate ship, fireboat, speedboat, lifeboat, airship, airliner, submarine, carrier ship and missile.

We then trained a robust 34 layer ConvNet neural network augmented with augmented training data procured using image transformations. The model was created on top of a pre-trained ResNet [12] architecture; using transfer learning and cross entropy loss optimization to achieve a validation accuracy of 93% with a batch size of 32, learning rate of 0.001 and 25 epochs.

### B. Creating Adversarial Examples

Adversarial perturbations are specially designed per-pixel values which when added to the image, seem imperceptible to the naked eye, but cause the image to be misclassified. Adversarial attacks can be broadly classified into white-box and black-box attacks. White-box attackers have access to the gradients from the training process of the underlying model, making it comparably easier to define their attack strategy. Black-box attacks on the other hand assume no knowledge of the model architecture or its underlying parameters. The attackers train a new model using their own images and labels; using it to construct adversarial examples to attack the victim's model with. Since white-box attacks have been traditionally harder to achieve good adversarial detection accuracy levels on, we will concentrate our attention over the course of this paper to white-box attacks.

Attacks can be targeted, or un-targeted. Un-targeted attacks do not attempt to change the classification to a specific target label. Instead, they just aim to change the classification to be different from the actual label. One of such methods called the Fast Gradient Sign Method (FGSM) [16]. It works by adding a pixel-wide perturbation on the signed gradient of the objective loss function on the true label with respect to the input image. From Eq. 1.0 for the perturbation function, J is the loss function for an output classification, y. The signed gradient of J with respect to the corresponding input image, x multiplied by a hyper-parameter ε (of range (0,1]) giving the function to calculate the perturbation needed to cause the maximum deviation from the true label.

$$\eta = \varepsilon * \text{sign}(\nabla_x J(\theta, x, y)) \text{ [16]} \qquad (1.0)$$

Epsilon (ε) values close to 1 would mean a more visible perturbation in the input image after noise addition. According to the results obtained on our implementation, we were able to reduce the overall accuracy on unseen data of 1162 images by ~49.5% using only a 0.05 value of epsilon in our attack. A perturbation with epsilon of 0.3 caused a complete disruption of the model with an accuracy drop of 75%. Fig. 1.1 depicts the rapid reduction of validation accuracy with a gradual increase in epsilon. We can clearly infer that by varying the multiplicative term - ε, most images that a human would have no problem classifying correctly would be egregiously misclassified by a deep learning model.

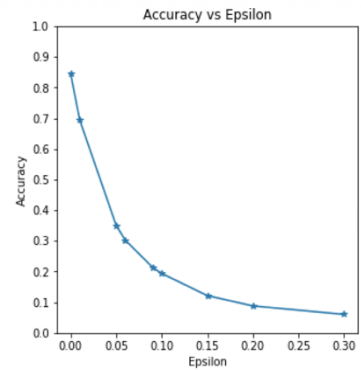

**Figure 1.1: Accuracy of the model with varying Epsilon values**

## VI. THE METAMORPHIC APPROACH

We apply metamorphic testing techniques in an attempt to identify if a given image is an unaltered or adversarial example. An adversarial example is assumed to be undetectable with the naked eye, but often causes the machine learning model to give the wrong prediction with a high degree of probability.

Our methodology is based on the idea that a machine learning model should be robust to small transformations in the input image, but the perturbation added to create an adversarial example may not survive the same transformation. Given an input image *I* which authenticity is unknown, we would like to verify that it is not an adversarial example with a simple test. We run the model with two versions of the image, *I* and *f(I)*, where *f* is a function that transform the image in some way. The model then produces two predictions, $y_1$ and $y_2$. If $y_1$ and $y_2$ are the same, we trust that the image was unaltered. If $y_1$ and $y_2$ differ, we conclude that the image was altered and the adversarial perturbation caused an unstable prediction.

The description above is somewhat simplified. We found that comparing just two outputs is not enough for a reliable detection. We run the transformation function several times with varying degrees of transformation and compare the outputs with the original. The comparison is not for equality. The output of the model is probabilistic, i.e. the probability that the input image belongs to a class. We measure how the output probability changes as the magnitude of the transformation increases. If the difference reaches a certain threshold, we conclude that the image is an adversarial example.

For this study, we considered various affine transformation functions. The goal was to identify a function that would cause little or no change in prediction for clean images, but a big change for adversarial images. Below we describe the transformations and how we apply them in more detail.

## A. Affine Transforms

Affine transformations are alterations on images that preserve both the collinearity relation between points and ratios of distance along the line [17]. This implies that all points on a straight line before the transformation remain so after transformations, while preserving their distance ratios.

$$\begin{bmatrix} x' \\ y' \\ 1 \end{bmatrix} = \begin{bmatrix} A & B & E \\ C & D & F \\ 0 & 0 & 1 \end{bmatrix} \begin{bmatrix} x \\ y \\ 1 \end{bmatrix} \quad (1.1)$$

$$\begin{bmatrix} x' \\ y' \end{bmatrix} = \begin{bmatrix} A & B \\ C & D \end{bmatrix} \begin{bmatrix} x \\ y \end{bmatrix} + \begin{bmatrix} E \\ F \end{bmatrix} \quad (1.2)$$

There are various types of affine transformations that linearly modify the spatial orientation of the images around an origin e.g. translation, scaling, reflection, rotation, shear.

Deep learning models are generally resistant towards small linear transformations in clean images without significant loss of classification accuracy. However, since adversarial perturbations over the images typify a directional gradient to increase the loss; we hypothesize that a linear transformation might cause a measurable difference in classification confidence. This would make it possible to classify adversarial images from clean images by seeing how the models respond to their linear transformations. We aimed to identify a cutoff that quantifies the separation between the range of confidence variations for clean and adversarial examples. We tested our hypothesis using four types of affine transformations.

## B. Detection of Adversarial Examples

We used a set of 1000 clean examples and their adversarial counterparts generated using FGSM with an epsilon value of 0.01. We purposely started with a relatively small set of 1000 adversarial/clean example pairs to see if the offset separation between clean and adversarial examples can be generalized without a large sample size.

While running transformation procedures, changes in behavior of adversarial and clean images are measured as follows: 1) Perform classification on the given image without transformation and capture the output label, $l_1$ and confidence value $v_1$ of that label. 2) Transform the image, re-run the classification and capture the output confidence value $v_2$ for label $l_1$. The change in behavior is measured by the degree of variation between $v_1$ and $v_2$.

We constrained our experiments to observing the variation in confidence/probability of the original output label $l_1$. For the purpose of this metric we consider that to be the "correct" label. It is important to note that we do not assume prior knowledge of the correct label. I.e. we do not try to detect if the classifier is good, just whether the input image is adversarial or clean.

### 1) Rotation Transform

Rotation is the linear transformation of a space around a point of origin. We gradually increased the angle of rotational transformation from 0.5 degrees with increments of 0.5 degrees to get groups of 60 sample sets. We kept the increments in angles small to observe possible deviation characteristics in the behavior of clean and adversarial examples to the transformation. We showed a visual confirmation of our hypothesis in Fig. 3. Using small rotation transformation (0.5 degrees), the adversarial examples have much greater variation in their "accurate label confidence" than the clean examples. The transformed clean example exhibits similar probability in its previously classified label but the transformed adversarial image sees a big drop in probability on the submarine label from ~73% to ~13%, with its output label changing from submarine to carrier ship.

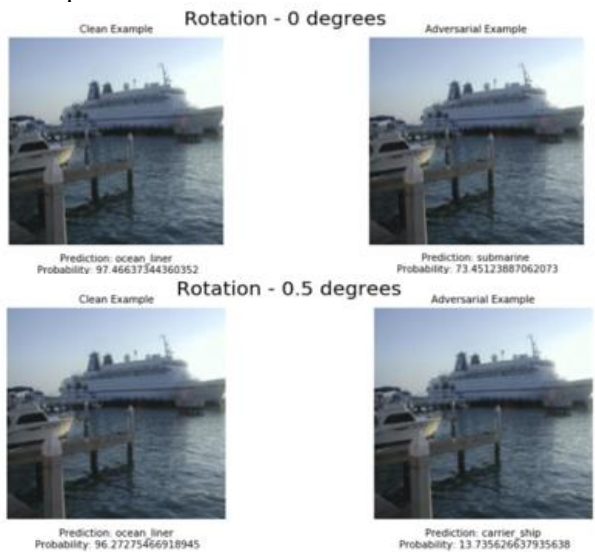

**Figure 2:** Example of model performance when rotation transformation was applied

Fig 4. Shows the variations in the classification accuracy of the correct label given the varying angles of rotation. We see a distinct drop in the label classification accuracy with incremental rotations of 0.5 degrees. The accuracy of clean images however seems to stay stable around the 100% mark. As mentioned before, we assume the "correct" label as the model's output on the unperturbed image. We tabulated the variation in the confidence levels towards the correct classification for clean and adversarial images at 0.5 degrees rotation in Table 1.

A graphical illustration of the trends between the mean and standard deviation of the label confidence values for the datasets within each of the 60 transformation groups is shown in Fig. 5. We can clearly see the stability in separation of the average values for the mean and standard deviation across our dataset. We decided to explore this behavior further and experiment if we could use the sum of mean and N times the standard deviation of clean examples as a cutoff to test for adversarial behavior. The N here is a hyper-parameter we can vary based on the distribution of image data. Furthermore, we observed that for some transformation angles, the variation in the confidence values between the adversarial example and its transformed image falls within the confidence variation range of the unperturbed/clean images. This results in lower accuracy when identifying adversarial examples. To reduce the effect of this anomalous case-based bias, we generalize the overall behavior by applying the deviation cutoff iteratively across multiple minutely spaced incremental rotations. We will henceforth refer to this method as "Hypothesis 1".

Using an unseen dataset of 2000 images, the adversarial detection accuracy using Hypothesis 1 across 60 iterations is shown in Fig. 6. We can see here that the detection accuracy has a near linear increase as the number of iterations of incremental rotation angle increase.

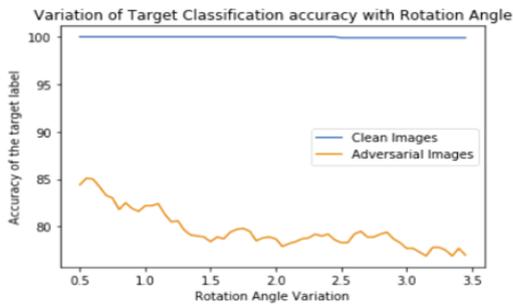

**Figure 3: Variations in the label classification accuracy**

**Table 1. Confidence values of clean and adversarial examples when using rotation of 0.5 degrees**

|      | CLEAN  | ADVERSARIAL |
|------|--------|-------------|
| MEAN | 0.484  | 32.935      |
| STD  | 2.117  | 28.305      |
| MIN  | 0      | 0           |
| 25%  | 0.002  | 5.672       |
| 50%  | 0.231  | 26.431      |
| 75%  | 0.206  | 57.970      |
| MAX  | 55.519 | 97.710      |

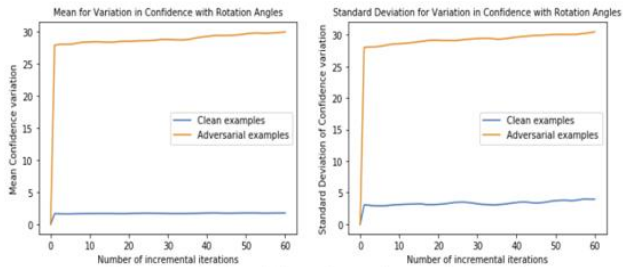

**Figure 4: Trends in the mean and STD of confidence values across different rotation angles**

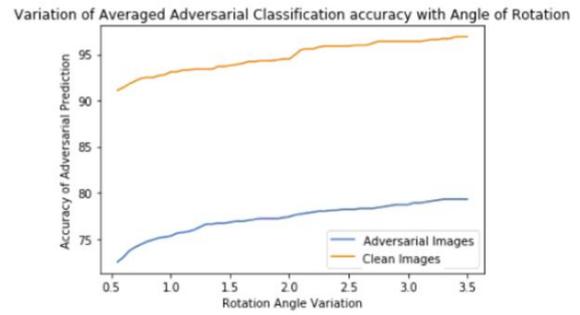

**Figure 6 -** After 60 iterations, we achieved 79.3% detection accuracy for adversarial examples and 97% for clean examples, with 88.1% overall accuracy.

*2) Shear Transform*

Image shear is the linear image transformation of every point to a fixed direction proportional to that of parallel lines in the plane. We again gradually increased the angle of shear transformation from 1 degree with increments of 0.9 degrees to get groups of 60 samples. Similar to our observations on rotation transformations, we see in Fig. 7 a much more pronounced drop in the target classification accuracy for adversarial samples with an increase in the angle of shear.

From Table 2, a change in shear of 1 degree causes a mean shift in confidence of 1.44 ± 2.75% in clean examples while the same transformation in their adversarial counterparts causes a mean confidence shift of 31.86 ± 26.7%. Fig. 8 shows the trends in the mean and the standard deviation of the confidence variations with each incremental angle increase for the adversarial and the clean images.

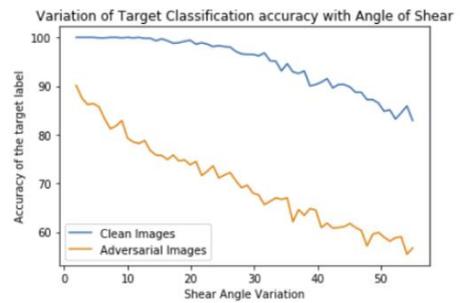

**Figure 7: Variations in the label classification accuracy**

**Table 2. Confidence values of clean and adversarial examples when using shear of 1 degree**

|      | CLEAN  | ADVERSARIAL |
|------|--------|-------------|
| MEAN | 1.437  | 31.863      |
| STD  | 2.750  | 26.704      |
| MIN  | 0      | 0           |
| 25%  | 0.033  | 5.383       |
| 50%  | 0.246  | 27.771      |
| 75%  | 1.249  | 55.164      |
| MAX  | 27.426 | 95.659      |

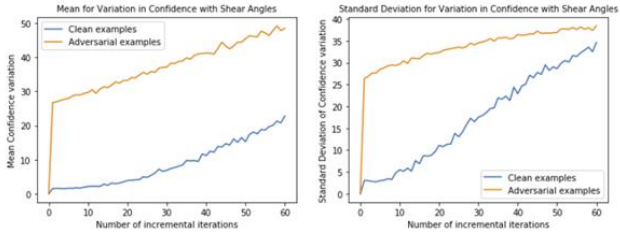

**Figure 8: Trends in the mean and STD of confidence values across different shear angles**

A clear demarcation in the mean values across the types of images should enable us to define clear cutoffs in separating the clean from adversarial examples. Towards higher shear values, the standard deviation average of clean examples averages out towards that of adversarial examples (using Hypothesis 1).

Using an unseen dataset of 2000 images, the increase in adversarial detection accuracy using our iterative transformation hypothesis across 60 iterations is shown in Fig. 9. We observe similar behavior to that using rotational transformation here.

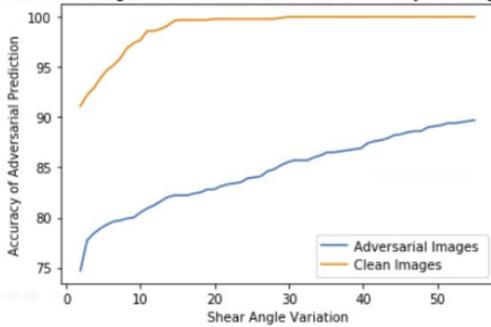

**Figure 9: After 60 iterations, we achieve an adversarial example detection accuracy of 90% and 100% for clean examples, with 95% overall accuracy.**

*3) Scale Transform*

Image zooming is the linear image transformation involving the geometric scaling up or down of image dimensions and cropping into the standard dimensions taken by the model. In this paper, we only explore scaling down of an image and gradually decrease the scale of each image from 1 in increments of 0.05 units over 60 zoom unit iterations.

Similar to our observations earlier, we see much sharper drops in the target classification accuracy for adversarial samples with increase in the negative scaling units. Fig. 10 shows a near parallel decrease across the two sets of examples, which should lead to a better accuracy in separating the two image types. Additionally, we observe that the scaling unit cutoff for our observations should lie somewhere around the 2.0 (20th iteration) mark since the separation between the two image types starts reducing at that point.

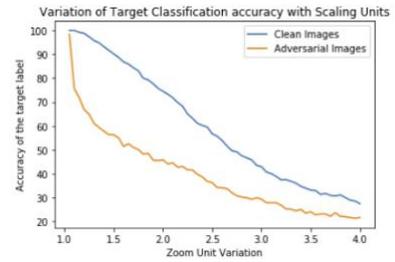

**Figure 10: Variations in the label classification accuracy**

In Table 3 we show the mean and standard deviation of the confidence variations at one incremental negative scaling unit. A change in scale of 1 unit causes a mean shift in confidence of $1.36 \pm 2.4\%$ in clean examples while the same transformation in its adversarial counterparts causes a mean confidence shift of $31.41 \pm 24.76\%$. Based on the observations regarding setting an earlier cutoff of scaling units, we use Fig. 11 to observe the averaged mean and standard deviations of our confidence variations with respect to iterations having incrementally increasing negative scaling units (maxed at 30 to give it a buffer over the 20th iteration). The average standard deviation values for our clean samples seem to take an unexpected dive upwards on higher scaling units, meaning that our prediction accuracy improvement on the final model should be optimized on a much earlier iteration when generalizing our cutoffs across different scaling units. Using Hypothesis 1 on unseen data, clearly the detection accuracy stabilizes at a much earlier iteration (Fig. 12).

**Table 3. Confidence values of clean and adversarial examples when using negative one scaling**

|        | CLEAN  | ADVERSARIAL |
|--------|--------|-------------|
| **MEAN** | 1.316  | 31.415      |
| **STD**  | 2.403  | 24.761      |
| **MIN**  | 0      | 0.001       |
| **25%**  | 0.034  | 6.561       |
| **50%**  | 0.224  | 29.443      |
| **75%**  | 1.205  | 53.333      |
| **MAX**  | 12.073 | 84.329      |

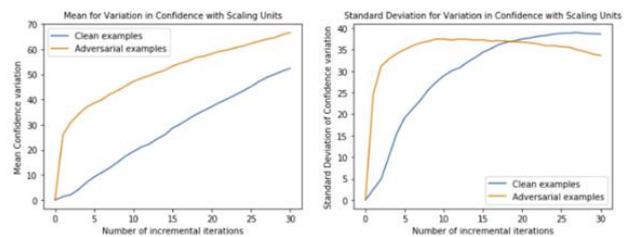

**Figure 11: Variations in the mean and STD of confidence values across different scaling units**

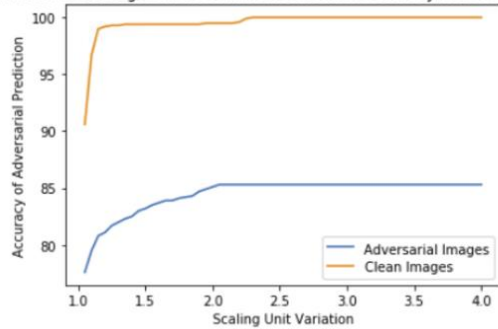

Figure 12: After 60 iterations, we achieve an adversarial example detection accuracy of 85.3% and 100% for clean examples, with 92.65% overall accuracy.

*4) Translate Transform*

The translate operation on images is a linear transformation that geometrically maps the position of each point to a new position using symmetric shifting. The image is translated both vertically and horizontally. The units of translation are fractions of the image width and height. We gradually increase the translation units from 0.05 incremented by 0.02 units per iteration.

From Fig. 13, we see an almost parallel correlation in the drops of target classification accuracy for adversarial and clean samples with translation across increasing units.

In Table 4, we show the average mean and standard deviations of the samples at an incremental translation unit change. A translation of 1 unit causes a mean shift in confidence of 2.33 ± 6% in clean examples while the same transformation in its adversarial counterparts causes a mean confidence shift of 37.25 ± 32.83%. Fig. 14 shows a clear differentiation between adversarial and clean examples in terms of their averaged observed variation for any given translation transformation unit used. Using hypothesis 1 on the unseen dataset, after 60 iterations (Fig 15), we achieve stellar results with an adversarial example detection accuracy of 93.7% and 100% for clean examples and an overall 96.85% accuracy. The results from testing our hypothesis over the four transformations are highly encouraging; being able to reach over 90% accuracy on an average without relying on a deep learning-based model. Additionally, we extracted quantifiable/differentiable behavior in adversarial examples using a dataset of only 1000 adversarial and clean images each.

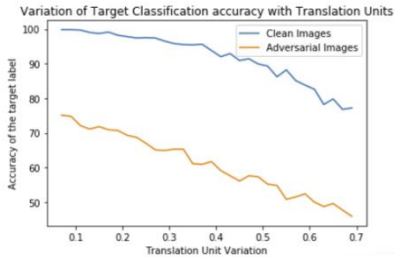

Figure 13: Variations in the label classification accuracy

Table 4. Confidence values of clean and adversarial examples for 1 translation unit

|  | CLEAN | ADVERSARIAL |
|---|---|---|
| MEAN | 2.330 | 37.255 |
| STD | 6.013 | 32.832 |
| MIN | 0 | 0 |
| 25% | 0.049 | 4.861 |
| 50% | 0.334 | 29.058 |
| 75% | 1.853 | 67.384 |
| MAX | 74.756 | 99.710 |

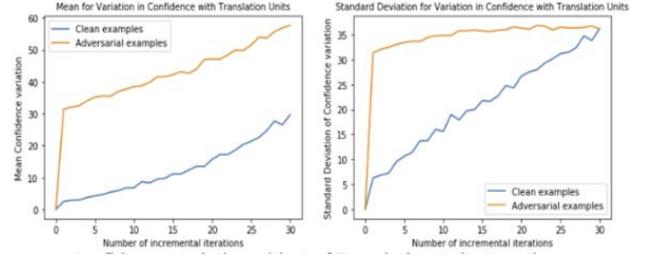

Figure 14: Variations in the mean and STD of confidence values

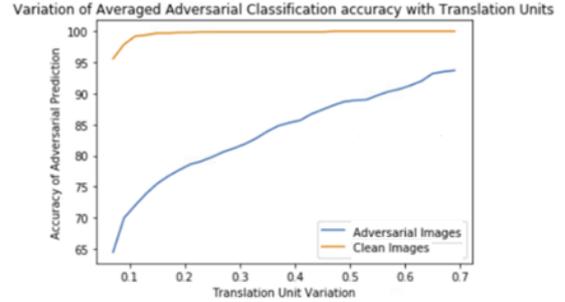

Figure 15: After 60 iterations, we achieve an adversarial example detection accuracy of 93.7% and 100% for clean examples, with 96.85% overall accuracy.

*C. Creating Metamorphic Relations*

Based on the positive results from our affine image transformation driven approach, we developed four metamorphic relations, see Table 5. We propose the following 4 metamorphic relations to classify examples as adversarial or clean images based on pre-defined probabilistic output relation axioms.

1. MR-1: Rotation transformation
2. MR-2: Shear transformation
3. MR-3: Scale Transformation
4. MR-4: Translate Transformation

Let *Xtrain* be the training data & *Xtest* be the test data with *F* acting as our 34 layer ResNet child model to classify features to their corresponding labels.

**MR-1: Rotation Transformation**

Iteratively finding confidence variation for $n$ samples for every $X$ across $N$ different rotational angles tested, we obtain the mean (say $M(n)$) and standard deviation (say $S(n)$) of the variation for clean examples in *Xtrain*. Given a test example $Xtest_n$, we iteratively test against whether the confidence variation for a rotational transform of $X$ degrees is greater than $M(n) + 1.5*S(n)$. The individual results across $N$ rotational angles are then grouped with an OR condition. If the output is 1, $Xtest_n$ is classified as an adversarial example with 79.3% confidence; else we classify it as a clean example with 97% confidence.

Table 5. The accuracy of various transformations for a given unseen image dataset

| Transformation | Clean Images | Adversarial Images | Average |
|---|---|---|---|
| Rotation | 97% | 79.3% | 88.15% |
| Shear | 100% | 90% | 95% |
| Scale | 100% | 85.3% | 92.65% |
| Translate | 100% | 93.7% | 96.85% |

**MR-2: Shear Transformation**

Iteratively finding confidence variation for $n$ samples for every $X$ across $N$ different shear angles tested, we obtain the mean (say $M(n)$) and standard deviation (say $S(n)$) of the variation for clean examples in *Xtrain*. Given a test example *Xtestn*, we iteratively test against whether the confidence variation for a shear transform of $X$ degrees is greater than $M(n) + 1.5*S(n)$. The individual results across $N$ shear angles are then grouped with an OR condition. If the output is 1, *Xtestn* is classified as an adversarial example with 90% confidence; else we classify it as a clean example with 100% confidence.

**MR-3: Scale Transformation**

Iteratively finding confidence variation for $n$ samples for every $X$ across $N$ different scaling units tested, we obtain the mean (say $M(n)$) and standard deviation (say $S(n)$) of the variation for clean examples in *Xtrain*. Given a test example *Xtestn*, we iteratively test against whether the confidence variation for a scaling of $X$ units is greater than $M(n) + 1.5*S(n)$. The individual results across $N$ scaling units are then grouped with an OR condition. If the output is 1, *Xtestn* is classified as an adversarial example with 85.3% confidence; else we classify it as a clean example with 100% confidence.

**MR-4: Translate Transformation**

Iteratively finding confidence variation for $n$ samples for every $X$ across $N$ different translation units tested, we obtain the mean (say $M(n)$) and standard deviation (say $S(n)$) of the variation for clean examples in *Xtrain*. Given a test example *Xtestn*, we iteratively test against whether the confidence variation for a translation of $X$ units is greater than $M(n) + 1.5*S(n)$. The individual results across $N$ translation units are then grouped with an OR condition. If the output is 1, *Xtestn* is classified as an adversarial example with 93.7% confidence; else we classify it as a clean example with 100% confidence.

*D. Results*

The metamorphic approach towards using properties of adversarial examples when transformed using the four relations seems to achieve high classification accuracy ranges bordering around 90%. Additionally, we observe that the shear and translate metamorphic relations are the most effective in detecting adversarial examples. We plan to study the effectiveness of using ensembles of affine transforms over future iterations of this paper. Additionally, our work opens doors towards considering affine transformations as a fast, lightweight and accurate way to detect adversarial examples over building complex multi-layer classifier models for the same. We plan to investigate the effectiveness of our method over adversarial images generated with other attack types like Carlini and Wagner [15], DeepFool [20], etc. We also plan to test our approach over datasets such as CIFAR [18], MNIST [19] to test the effectiveness of our method over lower resolution images.

VII. DISCUSSION AND FUTURE RESEARCH

In this paper, we have created a naval intrinsic deep learning model built using the transfer learning approach on the 34-layer ResNet architecture with 15 target labels. We then simulated adversarial attacks to disrupt our deep learning model's accuracy; to create our training data of 1000 adversarial/clean example pairs. We then studied the divergent characteristics in confidence/probability of the output label observed in adversarial images, when subjected to slight transformations in these images using affine linear transformations. This variation in the characteristics was quantified using the mean/standard deviation characteristics and subsequently generalized over an iterative algorithm comprising incremental transformation units. When tested on unseen validation data, our hypothesis showed very encouraging results. We used our successfully tested hypothesis to create four metamorphic testing axioms which are highly effective in filtering out adversarial samples. With our metamorphic approach in solving the problem, we were able to avoid using computationally intensive deep learning detector algorithms for our adversarial examples. In addition, we were able to achieve comparable/better results to the traditional deep learning detection approaches without having knowledge of the actual "correct label" or needing a large number of adversarial examples for training our models. Our devised metamorphic relations achieve a best-case detection accuracy of 96.85% when using translate transformation and gives encouraging results for the other three transformations too. We hope this study will inspire further exploration towards using metamorphic approaches in solving the adversarial detection problem plaguing the deep learning community; by studying their properties as a way of detecting and possibly reversing adversarial perturbations from different attack types.